# Deep Learning Based Steel Pipe Weld Defect Detection

Dingming Yang[a], Yanrong Cui[a]*, Zeyu Yu[b] and Hongqiang Yuan[c]

[a]*School of Computer Science, Yangtze University, Jingzhou 434023, China;*

[b]*School of Electronics & Information, Yangtze University, Jingzhou 434023, China;*

[c]*School of Urban Construction, Yangtze University, Jingzhou 434000, China;*

# Deep Learning Based Steel Pipe Weld Defect Detection

Steel pipes are widely used in high-risk and high-pressure scenarios such as oil, chemical, natural gas, shale gas, etc. If there is some defect in steel pipes, it will lead to serious adverse consequences. Applying object detection in the field of deep learning to pipe weld defect detection and identification can effectively improve inspection efficiency and promote the development of industrial automation. Most predecessors used traditional computer vision methods applied to detect defects of steel pipe weld seams. However, traditional computer vision methods rely on prior knowledge and can only detect defects with a single feature, so it is difficult to complete the task of multi-defect classification, while deep learning is end-to-end. In this paper, the state-of-the-art single-stage object detection algorithm YOLOv5 is proposed to be applied to the field of steel pipe weld defect detection, and compared with the two-stage representative object detection algorithm Faster R-CNN. The experimental results show that applying YOLOv5 to steel pipe weld defect detection can greatly improve the accuracy, complete the multi-classification task, and meet the criteria of real-time detection.

## Introduction

Steel pipes are widely used in high-risk and high-pressure scenarios such as oil, chemical, natural gas, shale gas, etc. If there is some defect in steel pipes, it will lead to serious adverse consequences. With the growing demand for steel pipe in China, more and more enterprises and even countries begin to pay attention to the quality and performance of steel pipe, and the defect detection and evaluation technology of steel pipe has become a research topic that researchers are keen on. At present, there are manual testing and X-ray testing. X-ray testing is one of the main methods for industrial non-destructive testing (NDT), and the test results have been used as an important basis for defect analysis and quality assessment of weld. X-ray detection can effectively detect the internal defects of steel pipe, but manual participation is still needed to determine the type and location of weld defects of steel pipe (Yun et al. 2009). Therefore, Applying object detection in the field of deep learning to the defect detection and identification of steel pipe welds can effectively improve the detection efficiency and promote the development of industrial automation.

With the wide application of artificial intelligence in the field of computer vision, machine learning and deep learning are widely used in object detection and image classification. Most predecessors used traditional computer vision methods to detect steel pipe weld defects (Yun et al. 2009; Wang et al. 2008; Malarvel et al. 2021). For example, Malarvel et al. (2021) used OSTU + MSVM-rbf (Multi–class Support Vector Machine) method to achieve multi-class detection of weld defects in X-ray images and achieved an accuracy of 95.23%. Nowadays, object detection algorithms based on deep learning are constantly developing, the recognition accuracy and detection time have been greatly improved compared with traditional computer vision methods. For example, Xiaojun Wu et al. (2021) used GAN (Generative Adversarial Network) to expand the insufficient defect data sets and proposed CFM (Coarse-to-Fine Module) to improve the segmentation algorithm with a good result; Yanqi Bao et al. (2021) proposed TGRNet (Triplet-Graph Reasoning Network) for metal generic surface defect segmentation and achieved good results. Previous studies have achieved good results, but there are also some shortcomings. Such as:

- Accuracy rate needs to be further improved;
- Different types of defects make it difficult to do multiple classifications with traditional computer vision methods;
- Detection time is too long to achieve real-time detection, so it is difficult to apply to the industrial field;

In view of the above problems, this paper applies the state-of-the-art YOLOv5 to the defect detection task of steel pipe weld.

## Materials and Methods

### *Profile of YOLOv5*

Joseph Redmon et al. (2016a) published YOLOv1 in 2015, which pioneered the single-stage object detection algorithm. This algorithm divides images into 7*7 grids, and each grid is responsible for the classification of objects and coordinate regression at the same time. Joseph Redmon et al. (2016b) published YOLO9000 in 2016 to make up for the shortcoming of YOLOv1 with fewer detection categories and low accuracy, but the detection of small targets is still poor. Joseph Redmon et al. (2018) published YOLOv3 in 2018, which draws on the idea of FPN (Tsung-Yi Lin et al. 2017), and solves the detection problem of small objects. Alexey Bochkovskiy et al. (2020) improved their algorithm by absorbing the tricks of various fields on the basis of the network structure of YOLOv3 and released YOLOv4, which greatly improved the detection efficiency and AP. Two months later, Ultralytics (a company) released YOLOv5 (Jocher et al. 2021).

According to the size of the model, YOLOv5 is divided into four versions: YOLOv5s, YOLOv5m, YOLOv5l and YOLOv5x. The larger the model is, the higher the accuracy will be, and the detection time of a single image will increase. Figure 1 shows the network structure of YOLOv5s. The technologies used in the Input of YOLOv5 include Mosaic data enhancement (Sangdoo et al. 2019), adaptive anchor calculation, and adaptive image scaling. The technology used in Backbone includes Focus structure and CSP structure. The techniques used in Neck include FPN+PAN structure; In Prediction, GIoU_Loss(Hamid Rezatofighi et al. 2019) is used to replace the ordinary IoU calculation method. YOLOv5 is slightly less capable than YOLOv4 in terms of performance, but much more flexible and faster than YOLOv4, so it has an advantage in model deployment.

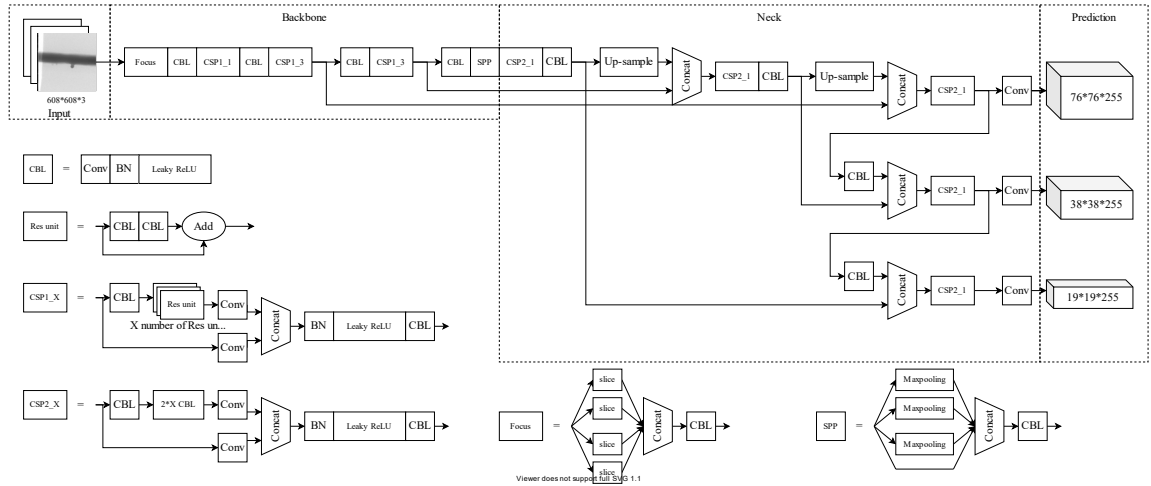

Figure 1. Network structure of YOLOv5s.

*Image acquisition device*

The real-time X-ray imaging system used in this paper is shown in Figure 2. The system mainly consists of welded pipe moving part, HS-XY-225 X-ray machine, PS1313DX high-speed digital panel detector, image capture card and display part. In the welded pipe moving part, the spiral submerged arc welded pipe is moved using a transmission vehicle with four longitudinal rollers fixed on the vehicle for rotating the spiral submerged arc welded pipe. The X-ray machine is fixed to the wall on one side and deep into the pipe on the other side, emitting X-rays that penetrate the weld seam. A flat panel detector absorbs the X-ray photons that pass through the weld, creating electronic data that retains information on the attenuation of the photons. An image capture card is used to convert the electronic data into a digital image sequence, which is then transferred to a computer for processing and display. Limited to hardware performance, only 8 X-ray images per second can be captured and processed.

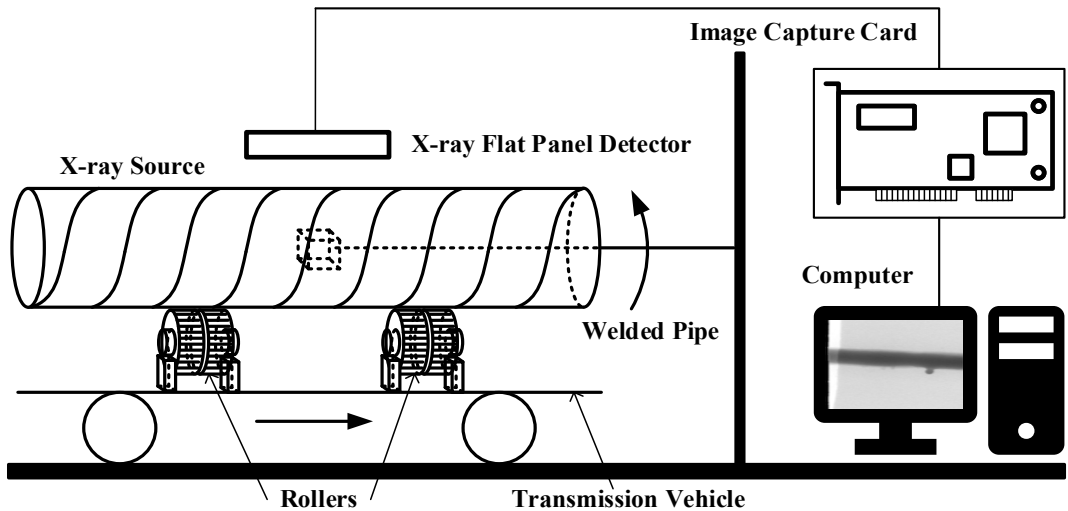

Figure 2. The real-time X-ray imaging system.

*Acquisition of dataset*

The raw video images of X-ray are provided by the cooperating factories in RAW format using the real-time X-ray imaging system in Figure 2. Through batch processing, the same width and height are cut out and exported as JPG images, and 3408 original images of weld defects of 8 types of steel pipe are obtained. Finally, Labelme (a software to mark object) was used to mark the defect area and defect category of steel pipe weld, which was then exported as the standard dataset format of YOLO or PASCAL VOC2007 (Ren et al. 2017). Figure 3 shows the types of steel pipe weld defects. The collected samples have a total of 8 types of defects, which are Blowhole, Undercut, Broken arc, Crack, Overlap, Slag inclusion, Lack of fusion, and Hollow bead. Table 1 shows the statistical table of steel pipe weld defects samples.

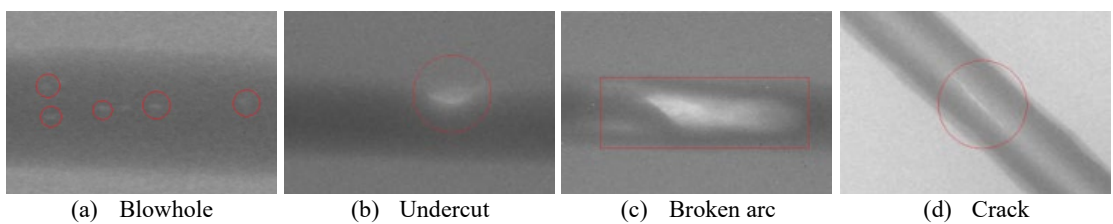

(a)　Blowhole　　　(b)　Undercut　　　(c)　Broken arc　　　(d)　Crack

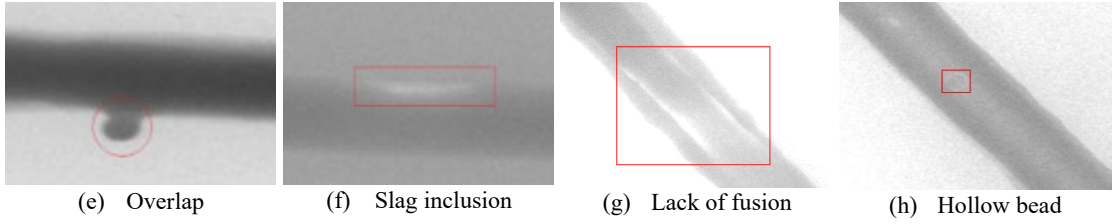

(e) Overlap     (f) Slag inclusion     (g) Lack of fusion     (h) Hollow bead

Figure 3. The example of steel pipe defects.

Table 1. Profile of sample images for 8 types of defects.

| Defect name | Number of original samples | Number of augmented samples | Label |
| --- | --- | --- | --- |
| Blowhole | 1339 | 12051 | blow-hole |
| Undercut | 35 | 315 | undercut |
| Broken arc | 531 | 4779 | broken-arc |
| Crack | 119 | 1071 | crack |
| Overlap | 219 | 1971 | overlap |
| Slag inclusion | 136 | 1224 | slag-inclusion |
| Lack of fusion | 416 | 3744 | lack-of-fusion |
| Hollow bead | 613 | 567 | hollow-bead |
| Totals | 3408 | 30672 | —— |

## *Data preprocessing*

### *Raw dataset analysis*

First of all, the original data should be analyzed so as to serve as a reference when setting parameters for deep learning and to accelerate the training speed. It can be seen from observation that X-ray pictures are black and white pictures, which can be converted into single-channel grayscale images. In this way, 2/3 pixels data can be compressed and the training speed will be accelerated. Then use Matplotlib (a python lib to draw diagram) to draw the scatter plot of the center point position of the bounding box and the length & width of the bounding box in turn to see if there are any extreme aspect ratios and abnormal data. As shown in Figure 4, it can be concluded that most bounding boxes are wider than their height and that the bounding boxes for cracked defects are close to a square. Secondly, the displacement of most defects is in the horizontal direction, and the displacement of Overlap defects is from the bottom right to the top left. The distribution of scatter is more even, and there are not many abnormal data.

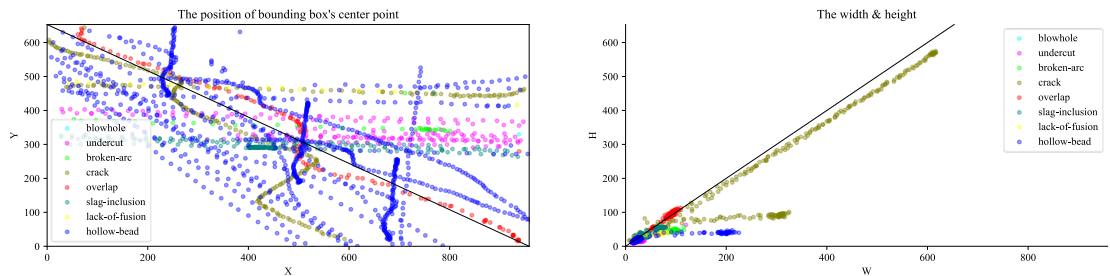

Figure 4. The analysis of original samples.

*Motion deblurring*

As shown in Figure 2, when the cylindrical steel pipe rotates in the assembly line, there will have relative movement between the X-ray camera used to film the weld defects of the steel pipe and the steel pipe in the direction of the weld. Moreover, the exposure time of the camera to shoot a single frame of weld defects is too long, so the motion blur will be generated. According to the research of Kupyn et al. (2018), motion blur will have an impact on the accuracy of object detection algorithm of YOLO series, so it is necessary to remove motion blur in some images. The process of motion deblurring is shown in Figure 5. First of all, we use the Hough Transform to detect the straight line at the weld edge. The direction of motion of the steel pipe can be estimated from the angle of the straight line (that is, the angle of image blur), and the distance of motion blur can be obtained from the frame rate of the camera and the speed of the steel pipe rotation. Then we used the estimated blurry kernel to deconvolution the original blurry image to get the result in Figure 5c.

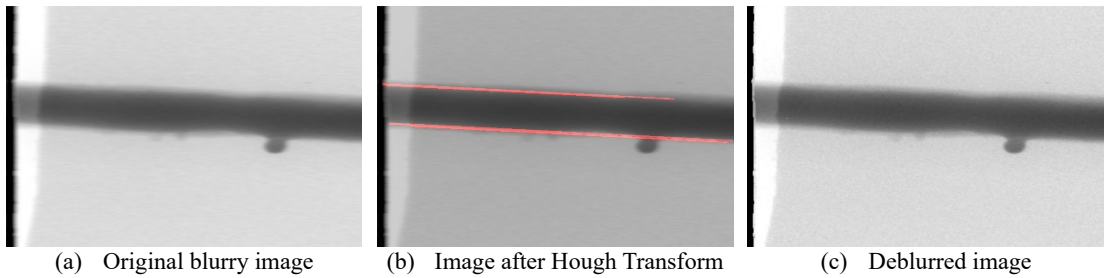

(a) Original blurry image    (b) Image after Hough Transform    (c) Deblurred image

Figure 5. The process of blind motion deblurring.

*Data enhancement*

Convolutional neural network (CNN) usually requires a large number of training samples to effectively extract image features and classify them. in order to effectively improve data quality and increase data feature diversity, the original data was enhanced to 9 times the original data by using light change, random rotation, random cut out, Gaussian noise addition, horizontal flipping, random adjustment of saturation, contrast and sharpness, random resize and random clipping. Thus, the over-fitting in the training stage is effectively reduced and the generalization ability of the network is improved. Figure 6 shows an example of what happens after data enhancement.

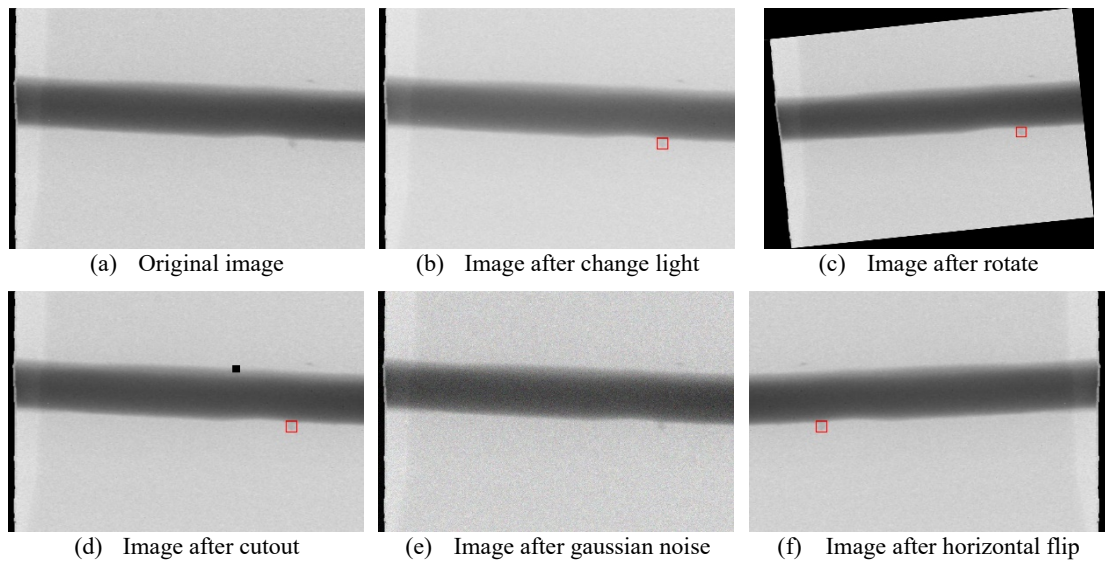

(a) Original image    (b) Image after change light    (c) Image after rotate

(d) Image after cutout    (e) Image after gaussian noise    (f) Image after horizontal flip

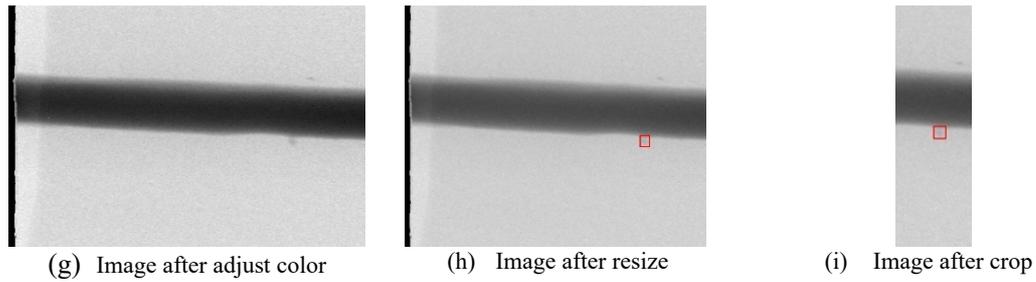

    (g)  Image after adjust color      (h)  Image after resize      (i)  Image after crop

Figure 6. The example after data augmentation.

## Experiments

### *Experimental environment*

Table 2 and Table 3 are the hardware environment and software environment of the experiment in this paper.

Table 2. The environment of hardware.

| Phase | CPU | GPU | RAM |
| --- | --- | --- | --- |
| Train | Intel(R) Xeon(R) CPU E5-2623 v4 @ 2.60GHz | Quadro P5000 | 30GB |
| Test | Intel(R) Core(TM) i7-4710MQ CPU @ 2.50GHz | GTX950M | 8GB |

Table 3. The environment of software.

| Phase | OS | Python | Model |
| --- | --- | --- | --- |
| Train | Linux-5.4.0-65-generic-x86_64-with-glibc2.10 | 3.8.5 | Official YOLOv5x |
| Test | Windows 10 professional edition | 3.8.0 | Official YOLOv5x |

### *Experimental process*

In this paper, the state-of-the-art deep learning algorithm YOLOv5 is used to train the detection model of steel pipe weld defects. After manually annotating the original image, the dataset is obtained through data enhancement, and then the dataset is converted into a single channel grayscale image. Because the dataset is relatively small, it is divided into training set and validation set in a ratio of 8:2. An experimental process designed in this paper is shown in Figure 7. After several epochs of YOLOv5 training, the training set and validation set obtained a model containing weight and bias parameters. In this paper, Recall, Precision, F1 score, mAP(mean of Average Precision) and detection time of single image were used as evaluation indexes.

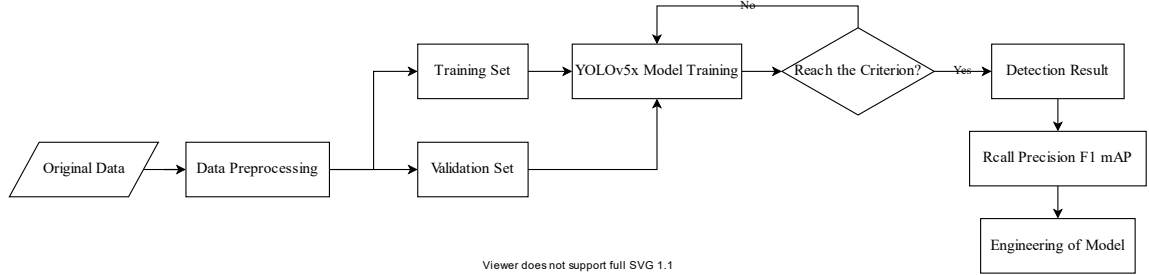

Figure 7. The flowchart of experiment.

The calculation method of Precision is shown in Formula (1). TP is the sample identified as true positive. In this paper, it is the identification of correct weld defects of steel pipe. FP is the sample identified as false positive. In this paper, it is the weld defect of steel pipe identified wrongly. The formula describes the proportion of true positive in the identified pictures of steel pipe weld defects. The calculation method of Recall is shown in Formula (2). FN is the sample identified as false negative, in this paper is the background of error identification; The formula describes the ratio of the number of correctly identified steel pipe weld defects to the number of all steel pipe weld defects in the dataset. The calculation method of F1 score is shown in Formula (3). When Precision and Recall are required to be high, F1 score can be used as an evaluation index. The calculation method of AP is shown in Formula (4). AP is introduced to solve the problem of limitation of Precision, Recall and F1 score single point value. In order to obtain an indicator that can reflect the global performance, In this paper, using the Interpolated average precision.

$$Precision = \frac{TP}{TP + FP} \qquad (1)$$

$$Recall = \frac{TP}{TP + FN} \qquad (2)$$

$$F_1 = \frac{2 * Precision * Recall}{Precision + Recall} \tag{3}$$

$$P_{\text{interp}}(k) = \max_{\hat{k} \geq k} P(\hat{k})$$
$$AP = \sum_{k=1}^{N} P_{\text{interp}(k)} \Delta r(k) \tag{4}$$

*Analysis of experimental results*

*Identify results and data analysis*

The detection result for 8 types of defects are shown in Figure 8. On the whole, both the position of defects and the confidence of classification are relatively good. Undercut's good performance in the case of a relatively small number of samples could not be attributed to the 8 data enhancement methods used in the data preprocessing stage of this paper and Mosaic data enhancement by YOLOv5. The Broken can still be identified as the same defect and obtain good confidence even if they are very different in appearance. Among them, the Slag inclusion defects are not obvious to distinguish from the background in the naked eye, and they are similar to the Undercut defects in appearance. Benefiting from repeated training, good results can also be achieved.

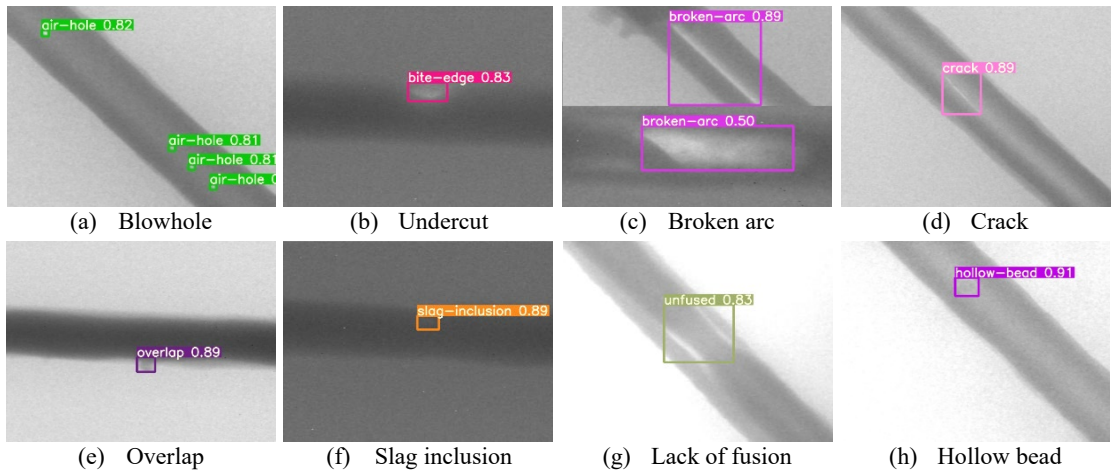

Figure 8. The result of detection.

As shown in Table 4, four evaluation indexes of each defect category in the last epoch are presented. On the whole, except for Blowhole defect, the accuracy of all other defects can be maintained between 0.962 and 1.00, the recall rate between 0.99 and 1.00, and the F1 score between 0.998 and 1.00. Blowhole defect due to its small defect target, a single steel pipe sometimes has dense pores, so the accuracy is lower than other types of defects. In the 218th epoch, the mAP of the model reached 99.02%, but after 633 epochs of training, the mAP decreased to 98.71%, showing some degree of over-fitting. The best training model saved in this paper can be used in the actual steel pipe weld defect detection and applied in the industrial production environment.

Table 4. Some statistical parameters of confusion matrix

| Type | blowhole | undercut | broken-arc | crack | overlap | slag-inclusion | lack-of-fusion | hollow-bead |
|---|---|---|---|---|---|---|---|---|
| Precision | 0.505 | 1.00 | 0.962 | 1.00 | 1.00 | 1.00 | 0.99 | 0.99 |
| Recall | 0.96 | 1.00 | 1.00 | 1.00 | 1.00 | 1.00 | 0.99 | 1.00 |
| F1 score | 0.661 | 1.00 | 0.98 | 1.00 | 1.00 | 1.00 | 0.99 | 0.994 |
| AP | 0.951 | 0.995 | 0.992 | 0.995 | 0.995 | 0.995 | 0.978 | 0.995 |
| mAP@0.5 | | | | 0.987 | | | | |

*Performance comparison of weld defect detection algorithm for steel pipe*

As shown in Figure 9, we used the same dataset to conduct experiments respectively in Faster R-CNN (Ren et al. 2017; Bubbliiiing 2020) and YOLOv5 (Jocher et al. 2021), then compared the precision data and total loss data generated during the experiment. As shown in Figure 9a, Faster R-CNN calculates the precision mean after each epoch of training, and has a tendency of descending and then slowly climbing, with unstable values in the second half. YOLOv5, on the other hand, started off with a shaky precision, then slowly climbed up and settled down. As shown in Figure 9b, the total loss of Faster R-CNN tended to be stable between 50-100 epoch, and then had two relatively large wave peaks. Since Faster R-CNN uses the Adam (Diederik Kingma et al. 2014) optimizer, it can converge faster than SGD (Stochastic Gradient Descent). The initial total loss of YOLOv5 was relatively small and tended to be stable between 100-150 epoch, with a small peak around 160 epoch. YOLOv5 also uses the optimizer Adam, and the initial value of Momentum is 0.999. In general, compared with the Faster R-CNN, YOLOv5 has better convergence speed in precision & total loss and stability after convergence than the Faster R-CNN.

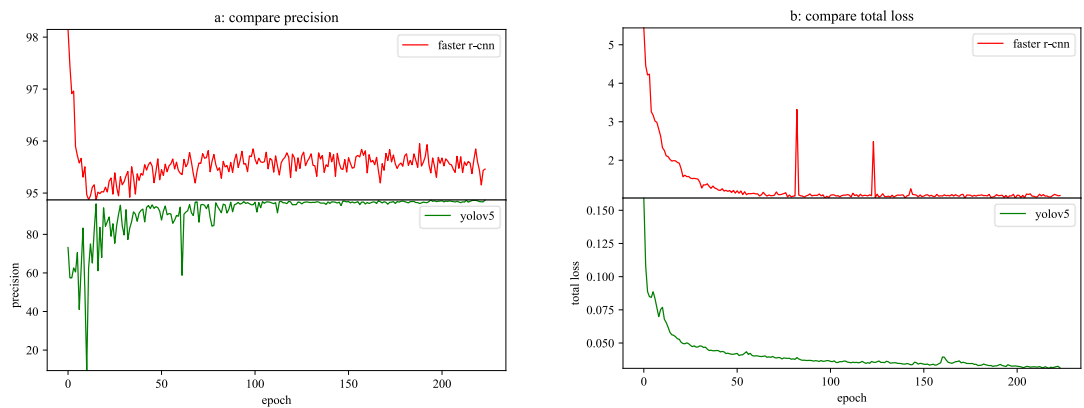

Figure 9. Compare with Faster R-CNN

As shown in Table 5, a comparison is made between GAN+CFM, OSTU+SVM, Faster R-CNN+ResNet50 and YOLOv5. On the whole, the defect detection algorithm based on deep learning is better than the defect detection algorithm based on traditional computer vision in both performance and detection time of a single image. Among them, GAN+CFM algorithm takes the longest time; OSTU+MSVM-rbf algorithm has the lowest accuracy. YOLOv5 is superior to Faster R-CNN in both accuracy and detection time of a single image. The detection time of a single image satisfies the engineering work of the model in the later stage of this paper. YOLOv5's detection speed is to be expected because it's one-stage. Another kind of object detection algorithms is two-stage. For example, the Faster R-CNN algorithm forms region proposals (which may contain areas of the object) first and then classifies each region proposal (also corrects the position at mean time). This type of algorithm is relatively slow because it requires multiple runs of the detection and classification process.

Table 5. Performance comparison of steel pipe defection algorithms

| Object detection model | Accuracy or Precision/% | Detection time per picture/s |
| --- | --- | --- |
| GAN+CFM (Wu et al. 2021) | 85.9 acc(mIoU) | 0.132 |
| OSTU+MSVM-rbf (Malarvel et al. 2021) | 95.23 acc | —— |
| Faster R-CNN+ResNet50 (Ren et al. 2017) | 95.5 acc(mAP@0.5=78.1) | 0.437 |
| YOLOv5x (Jocher et al. 2021) | 97.8 pre(mAP@0.5=98.7) | 0.120 |

**Conclusion**

In the field of steel pipe weld defect detection, deep learning method has more advantages than traditional computer vision method. Convolutional neural network does not need to extract image features manually, and can realize end-to-end input detection and classified output. The research of this paper has the following three contributions:

- Applying the state-of-the-art object detection algorithm YOLOv5 to the field of steel pipe weld defects detection, The detection accuracy of steel pipe weld defects and the detection time of a single image are pushed to a new height level, with the accuracy reaching 97.8% (mAP@0.5=98.7%). Under the YOLOv5x model testing, the detection time of a single picture is 0.12s (GPU=GTX950M), which meets the real-time detection on the steel pipe production line;
- Did a lot of work in the data preprocessing stage, combining the traditional data enhancement method with the Mosaic data enhancement method of YOLOv5, which not only greatly increased the size of the dataset, but also effectively reduced the over-fitting of the training;

- The results of YOLOv5 were compared with previous defect detection algorithms, and the advantages of YOLOv5 in model deployment and model engineering were demonstrated on the premise of comprehensive indicators.

This study can provide methods and ideas for real-time automatic detection of weld defects of steel pipe in industrial production environment, and lay a foundation for industrial automation. Although this paper uses state-of-the-art deep learning algorithm and convolutional neural network model for real-time detection of steel pipe weld defects in industrial production scenarios, its performance and performance are also relatively good. However, in the case of limited dataset, other defects which not in the dataset cannot be correctly identified. In this case, we can use traditional computer vision or mathematical methods to build an expert system to identify other defects that do not appear in the dataset. It is also possible to design an automatic updating model system in combination with Few-shot learning in engineering, which is used to manually label the type and bounding box coordinate information by the quality inspector when the defect cannot be identified, so that the system can automatically learn and update the model. These deficiencies point out the direction and provide ideas for the follow-up research.

**Author Contributions**